\documentclass[10pt,twocolumn,letterpaper]{article}

\usepackage{cvpr}              

\usepackage{graphicx}
\usepackage{amsmath}
\usepackage{amssymb}
\usepackage{booktabs}

\usepackage[pagebackref,breaklinks,colorlinks]{hyperref}

\usepackage[capitalize]{cleveref}
\crefname{section}{Sec.}{Secs.}
\Crefname{section}{Section}{Sections}
\Crefname{table}{Table}{Tables}
\crefname{table}{Tab.}{Tabs.}

\def\cvprPaperID{*****} 
\def\confName{CVPR}
\def\confYear{2023}

\begin{document}

\title{Linking vision and motion for self-supervised object-centric perception}

\author{Kaylene Stocking\textsuperscript{$1^*$,2}, Zak Murez\textsuperscript{1}, Vijay Badrinarayanan\textsuperscript{1}, Jamie Shotton\textsuperscript{1}, Alex Kendall\textsuperscript{1},\\ Claire Tomlin\textsuperscript{2}, Christopher P. Burgess\textsuperscript{$1^*$} \\
\textsuperscript{1}Wayve \;\;\; \textsuperscript{2}UC Berkeley \\
{\tt\small \{kaylene, tomlin\}@berkeley.edu, research@wayve.ai, research@chrisburgess.me.uk}
}
\maketitle

\begin{abstract}
Object-centric representations enable autonomous driving algorithms to reason about interactions between many independent agents and scene features. Traditionally these representations have been obtained via supervised learning, but this decouples perception from the downstream driving task and could harm generalization. In this work we adapt a self-supervised object-centric vision model to perform object decomposition using only RGB video and the pose of the vehicle as inputs. We demonstrate that our method obtains promising results on the Waymo Open perception dataset. While object mask quality lags behind supervised methods or alternatives that use more privileged information, we find that our model is capable of learning a representation that fuses multiple camera viewpoints over time and successfully tracks many vehicles and pedestrians in the dataset. Code for our model is available at \tt{https://github.com/wayveai/SOCS}.
\end{abstract}

\let\thefootnote\relax\footnote{$1^*$ Work carried out at Wayve}
\section{Introduction}

People and robots both operate in environments where objects - things that tend to behave as single coherent entities - are a natural organizing principle for perception. Objects play a central role in human vision. We group features into objects \cite{wolfe2015sensation}, describe our surroundings in terms of them, and seek out semantic labels for ones we're unfamiliar with \cite{gopnik1997words}. When using visual representations for downstream tasks such as robotics, object-centric models are therefore desirable if only because they are easier for people to interpret than end-to-end models, which is important for verifying safety and engendering trust in vision-enabled systems. More than this, however, object-centric representations enable a diverse and powerful set of tools for reasoning about the world, such as models for physics understanding \cite{wu_slotformer_2023}, multi-agent prediction and planning \cite{van2011reciprocal}, and causal reasoning \cite{vowels2022dags}. Representations that support these kinds of models are likely to be crucial for autonomous driving, where choosing the best action requires reasoning about many interacting agents as well as physical factors.

Traditionally, object-centric representations have been obtained by training supervised object detection models and extracting object properties such as position and velocity from their predictions. This approach has two major drawbacks. First, it requires an annotated dataset that matches the desired detections, which is expensive to obtain at scale and may introduce unwanted bias. To make the system work with new kinds of objects or in visually novel situations, new annotations have to be collected. Not having enough of the right kind of labeled data can hamper generalization, which promises to be one of the key advantages of object-centric models \cite{greff2020binding}. 

Second, creating object representations from the predictions of supervised vision models introduces a decoupling between the perception and downstream decision-making components of the system. For example, should a person riding a bike be treated as one object or two? What about two people on a tandem bike? Is it important that they're waving at a pedestrian on the sidewalk? What if they're making a hand signal for turning? The right answer to these questions depends on how the information will be used to make decisions. Ideally, success at using perception for action should feed back into improving perception, leveraging the power of end-to-end learning to find better object-centric features than what can be designed by hand.

These considerations motivate \emph{self-supervised} object-centric perception, where the model learns to encode images into a latent space that divides up relevant information about the scene into a number of discrete ``slots." The contents of these slots can then be decoded into a self-supervised target such as reconstructing the original RGB input, as well as for downstream tasks. Various methods have been reported for encouraging models to encode information about different objects in separate slots, such as by using an encoder that has slots compete for attention over pixels \cite{locatello2020object} or incentivizing slot disentanglement with a variational autoencoding loss \cite{kabra_simone_2021}. However, these methods have struggled to obtain good results with complex real-world data. Recently, the algorithm SAVi++ has shown segmentation on the Waymo Open dataset of real-world driving videos \cite{kipf_conditional_2022}. However, these results required additional depth supervision, as well as initializing the slots with bounding boxes around the desired objects to be tracked for best performance.

In this work we present a model which uses only RGB video and information about camera movement to obtain promising self-supervised segmentation results on real-world driving videos. Camera movement can be readily obtained in an autonomous vehicle setting (for example from simultaneous localization and mapping (SLAM) or wheel odometry), requiring no specialized sensors such as lidar. This makes it a particularly economical form of auxiliary input. Our model learns to segment many vehicles and pedestrians from the background and additionally shows consistent tracking of objects over time and across multiple cameras.

\section{Related work}

Object-centric representation learning in the self-supervised setting has received increased attention in recent years. A survey of learning and using these representations can be found in \cite{greff2020binding}. Techniques that operate on a single image work well with very simple synthetic scenes \cite{locatello2020object,greff2019multi}, but for more complex scenes the correct segmentation is ill-defined. Sequences of multiple images over time and/or space provide richer information about 3D shape and allow decomposition of more complex scenes \cite{kipf_conditional_2022,kabra_simone_2021,sajjadi2022object}. However, most approaches have still been demonstrated only on synthetic datasets, with segmentation of real-world images remaining an open challenge.

The recent SAVi++ model \cite{elsayed2022savi++} has shown promising instance segmentation and tracking of vehicles in the real-world Waymo Open dataset. However, this model requires ground-truth depth images in addition to RGB image input, and strong performance additionally requires that bounding boxes for the objects to be tracked to be provided at test-time.

Another recent model, STEVE, does not use additional supervision and presented results on a dataset of real videos of aquariums and traffic from a fixed overhead camera \cite{singh_simple_2022}. However, the fixed background in these videos simplifies the task. Additionally, the freeway environment in the traffic videos has less variety in background textures and shapes than driving through different environments including busy city streets.

The Object Scene Representation Transformer (OSRT) has a similar architecture to our proposed model, but uses novel viewpoint synthesis, rather than autoencoding, as the training task \cite{sajjadi2022object}. Although it was originally only demonstrated on synthetic datasets, recent work has used used OSRT as the vision backbone in a real-world robotic manipulation task, albeit in a simple environment consisting of a blank table with a small number of solid-colored shapes \cite{driess2023palm}.

The closest work to ours is SIMONe \cite{kabra_simone_2021}, which uses variational autoencoding and independent per-slot reconstruction to learn an object-centric decomposition of image sequences. To adapt SIMONe for complex real-world driving videos, we: 
\begin{enumerate}
    \item Provide the ground-truth transform matrix for each input image instead of learning self-supervised ``frame latents,"
    \item Change the reconstruction decoder from a unimodal gaussian distribution to a gaussian mixture distribution, as described in section \ref{sec:decoding_distribution}.
\end{enumerate}
Additionally, we experiment with using multiple camera viewpoints and adding prediction of the future ego-vehicle pose as an auxiliary training task. This task can be viewed as behavioral cloning, which is a common technique in end-to-end autonomous driving \cite{chen_learning_2022}. Previous work has demonstrated the potential of using an object-centric representation for predictions involving multiple dynamic objects \cite{wu_slotformer_2023}.

\section{Approach}
\begin{figure*}
    \centering
    \includegraphics[scale=0.36]{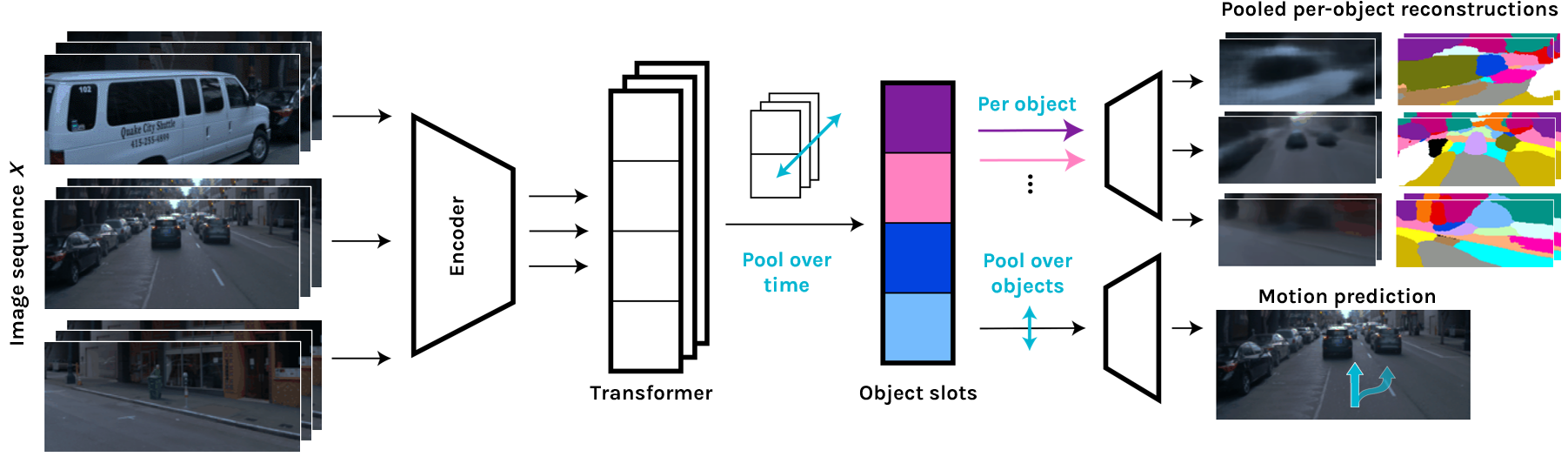}
    \caption{An overview of the proposed model architecture. From left to right, a sequence of images (optionally with multiple camera viewpoints such as front-left, front-forward, and front-right) is encoded into a set of feature patches with a CNN and decomposed into a set of $K$ slots via a transformer. In the top right, the slots are independently decoded into predicted RGB value distributions for each pixel. In the bottom right, the slots are pooled and used for auxiliary tasks such as predicting the future trajectory of the ego vehicle. The training losses described in section \ref{sec:losses} encourage the model to store information about different objects in each slot.}
    \label{fig:architecture}
\end{figure*}

For our model architecture, we build on the view-supervised variant of SIMONe \cite{kabra_simone_2021}. The goal of the model is to decompose the scene into a set of $K$ \emph{object slots} that encode information about each object in the scene. These slots are obtained as follows: first, the input $X$ (a sequence of $F$ images, optionally from multiple camera viewpoints), is processed in parallel by a standard convolutional neural network (CNN), resulting in a set of feature patches. The patches are concatenated with positional embeddings that encode the position of each patch within its source image, as well as the time and viewpoint transform matrix associated with the source image. They then serve as input tokens to a decoder-only transformer. The output tokens are averaged across the image dimension. In the original SIMONe model this dimension corresponds to images from a single camera captured over several timepoints, but in this work we use three different camera viewpoints, and the average operation is over both time and viewpoint. Finally, each token is decoded via an MLP into $m$-dimensional vectors $z_k$ and $\sigma_k$ that contain, respectively, the predicted mean and variance of the latent information in a single slot. If the number of input tokens is different from the desired number of object slots, an optional spatial pooling operation across the token features is added halfway through the transformer layers.

\subsection{Training and losses}
\label{sec:losses}
In order to encourage the model to store information about different objects in different slots, during training we apply three complementary losses. The first is the KL-divergence between each slot latent vector and a unit normal distribution, summed over all slots: 

\begin{equation}
    \mathcal{L}_{KL}(X) = \sum_k D_{KL}(q_k(X) || p(\cdot))
\end{equation}
where $X$ are the input frames, $q_k(X)=\mathcal{N}(z_k(X), \sigma_k(X)\mathbb{I})$ is an $m$-dimensional normal distribution with mean and variance predicted by the model for slot $k$, and the prior distribution $p(\cdot)$ is a unit spherical normal distribution. Intuitively, this loss encourages the model to avoid representing the same object using multiple slots, as doing so would incur a larger penalty than using a single slot to represent the object and letting the others remain closer to the unit normal distribution. This loss also encourages disentanglement between the features represented in each dimension of the latent vector.

The second loss is based on the model's ability to perform an object-wise reconstruction task. First, object latent vectors $o_k \sim q_k(X)$ are sampled from the latent distribution for each slot $k$. Then, each $o_k$ is independently decoded into pixel-wise predictions using a spatial broadcast decoder \cite{watters2019spatial}. To keep the computational and memory requirements of the model manageable, only a random sample of $N$ pixels is decoded from each input sequence during training. The per-slot predictions for each pixel parameterize a distribution over possible RGB values, $p(x^{(n)} | o_k)$, and a logit $\hat{\alpha}_k$ that (after normalization across slots) represents the likelihood that slot $k$ represents the pixel. To obtain the final prediction for each pixel, we take a weighted mixture of each slot's predictions:

\begin{align}
    \label{eqn:slot_mixture}
    p(x^{(n)} | o_1, ..., o_K) &= \frac{1}{K} \sum_k \Bar{{\alpha}}_k [p(x^{(n)} | o_k)] \\
    \Bar{\alpha}_k &= \frac{\text{exp}(\hat{\alpha}_k)}{\sum_{k'} \text{exp}(\hat{\alpha}_{k'})}
\end{align}
where the per-slot distributions for pixel $x^{(n)}$ are weighted by the softmaxed $\hat{\alpha}_k$ values. The distribution of $p(x^{(n)} | o_k)$ is discussed in more detail in section \ref{sec:decoding_distribution}.

The final reconstruction loss is the log probability of the ground truth RGB values for each pixel under the mixture distribution:
\begin{align}
    \mathcal{L}_{recon}(X) =& \frac{-1}{N} \sum_n \text{log} p(x^{(n)} | o_1, ..., o_K) \\
    & o_k \sim q_k(X)
\end{align}
Since the object latent vectors from each slot are decoded independently, the model is forced to use only information encoded in a single slot at a time when predicting the RGB values for each pixel. Therefore, intuitively this loss encourages the model to store all of the information it needs to predict the color of pixel $n$ in a single slot.

Optionally, the learned slot representations can also be used for auxiliary tasks. In this work, inspired by the close coupling between objects that matter in the autonomous driving context and representations that are useful for predicting good driving actions, we experiment with predicting the future waypoints of the ego vehicle as the auxiliary task. After the image-wise pooling step, the slot tokens are passed through two transformer decoder layers, averaged, and decoded into a series of predicted offsets, $\hat{s}$, in the $xy$-plane of the ego reference frame via a single-layer MLP. We then apply the following task loss:
\begin{equation}
    \mathcal{L}_{task}(s) = \sum_{t'} ||s_{t'} - \hat{s}_{t'}||_1
\end{equation}
where the summation is over each future time point. We use 16 future waypoints in the ego reference frame at a frequency of 10 Hz, starting $t_0'=0.1s$ after the final image frame. 

The final training loss is similar to the negative-ELBO loss in variational autoencoding \cite{higgins2017beta} with the addition of the auxiliary task loss:
\begin{equation}
    \mathcal{L}_{total}(X,s) = \mathcal{L}_{recon}(X) + \omega_{task}\mathcal{L}_{task}(s) + \beta \mathcal{L}_{KL}(X)
\end{equation}
where the $\beta$ and $\omega_{task}$ hyperparameters balance the different loss terms.

\subsection{Additional Model outputs}
In addition to parameterizing the weighted mixture pixel distribution, the  $\Bar{\alpha}_k$ weights serve as per-slot alpha masks that make it easy to see which pixels are best represented by each slot. Taking the argmax of this across slots results in a predicted segmentation of the scene. This segmentation can aid in model debugging and interpretation. For example, failing to track a particular vehicle with a mask indicates that the model is not disentangling that object's features from other features in the scene, and therefore is not independently representing its dynamics.

The object slots or latent vectors may also be decoded into other outputs besides image reconstruction or waypoint prediction. Other potential downstream tasks could include video prediction \cite{wu_slotformer_2023}, production system models \cite{alias2021neural}, or action-conditioned world models \cite{sun_smart_2022}. Which auxiliary tasks improve performance synergistically in end-to-end robot learning is an exciting open question.

\subsection{Slot decoding distribution}
\label{sec:decoding_distribution}

The original SIMONe model uses a unimodal normal distribution for the predicted pixel RGB values $p(x^{(n)} | o_k)$. (Note that throughout this section we refer to an RGB tuple as a single mode, but in reality the R, G, and B channels are treated independently.) We find that this distribution causes the model to over-rely on differences in color when decomposing the scene into object slots. This leads to failure cases such as separately segmenting the body and windshield of vehicles, and failing to pick out objects with similar colors to the background. We speculate that this is due to the fact that around the border of two differently-colored regions of a single object, the model may be uncertain about which color to assign a given pixel. To express this uncertainty with a unimodal color distribution, the model is forced to assign the different colored regions to different slots, and use the per-slot $\hat{\alpha}_k$ weights to give likelihoods to each color.

For our architecture, we replace the unimodal distribution in SIMONe with a multi-headed normal distribution to alleviate this issue. Qualitatively, we find that this leads to segmentations that better reflect the motion of objects. For each pixel and each slot $k$, the decoder outputs $H$ modes with a predicted mean RGB tuple $\hat{\mu}_k^{(h)}$ and $\alpha_k^{(h)}$ logit which determines the weight of each mode. (Note that there is additionally a separate ``global" $\hat{\alpha}_k$ logit which controls the $k$th slot's contribution to the total mixture distribution, as shown in equation \ref{eqn:slot_mixture}.) The per-slot distribution is therefore
\begin{equation}
    p(x^{(n)} | o_k) \sim \frac{1}{H} \sum_h \frac{\text{exp}(\alpha_k^{(h)})}{\sum_{h'} \text{exp}(\alpha_k^{(h')})} \mathcal{N}(\hat{\mu}_k^{(h)}, \sigma_x)
\end{equation}
where the variance of the normal distribution, $\sigma_x$, is a hyperparameter. When $H=1$, this simplifies into the decoding distribution in SIMONe:
\begin{equation}
    p(x^{(n)} | o_k) = \mathcal{N}(\hat{\mu}_k, \sigma_x)
\end{equation}
In our experiments we use $H=3$ and $\sigma_x=0.08$.

\section{Results}
We evaluate our approach on the Waymo Open Perception dataset \cite{Sun_2020_CVPR}. We train our model on sequences of 8 frames at 5 Hz, with input images cropped to remove the sky and resized to a resolution of 96 x 224. The validation dataset includes approximately 2000 frames with semantic and instance segmentation labels that we leverage to evaluate our model's ability to segment and track objects. Of these frames, we sample 208 sequences, which like the training data are 8 frames at 5 Hz, and use the front-left, front-forward, and front-right camera viewpoints for a total of 24 frames per sequence. 

\subsection{Qualitative results}
\begin{figure}
    \includegraphics[scale=0.3]{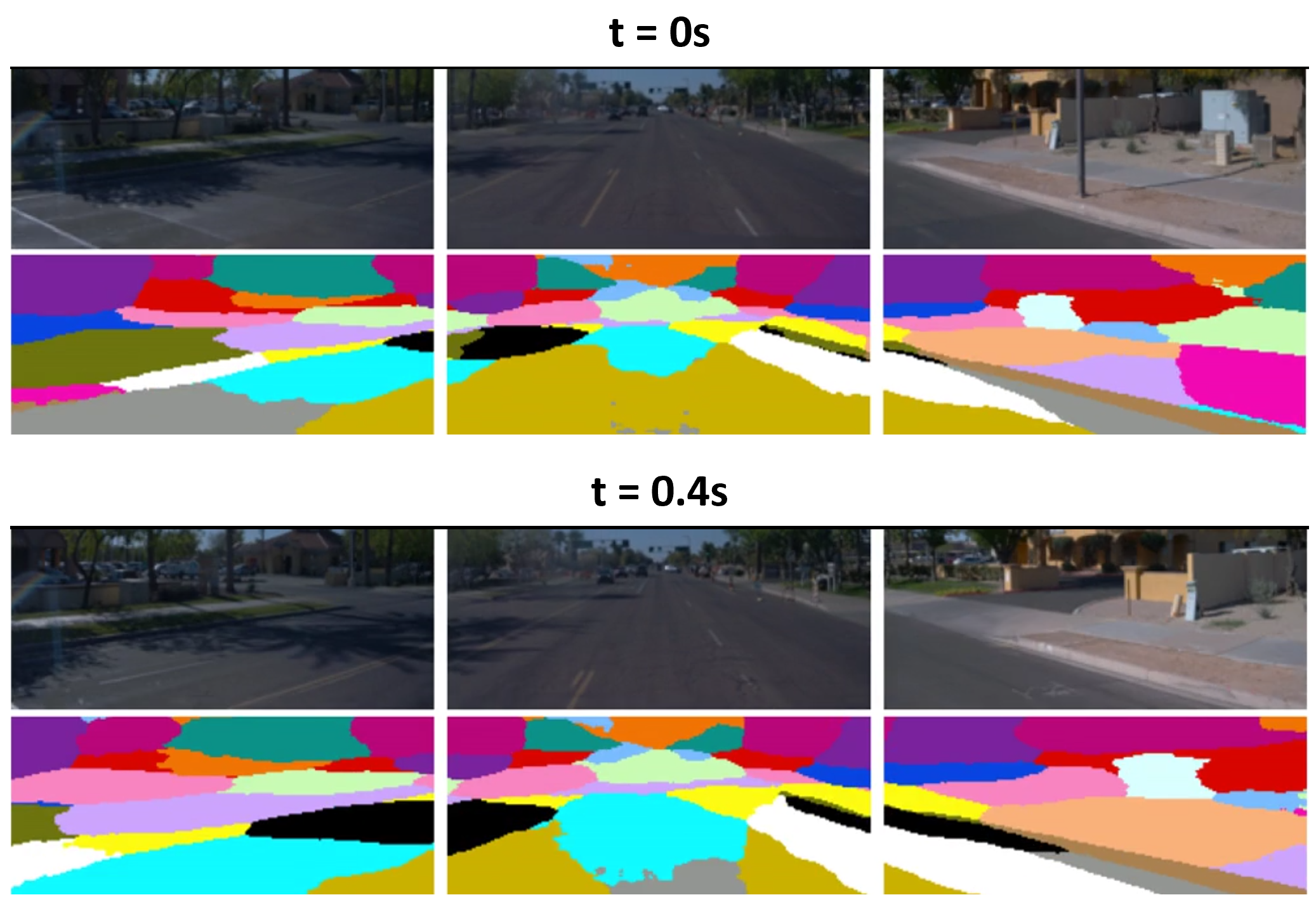}
    \caption{Our model partitions the scene between objects in a way that is consistent across time and multiple camera viewpoints, despite being given no explicit fusion information besides the camera transformation matrix for each input image. For example, in this scene note the tree shadows on the pavement in the left camera image and the orange wall on the right. The segmentation masks in this figure are derived from the slot with the largest predicted weight for each pixel.}
    \label{fig:fusion}
\end{figure}

\begin{figure*}
    \centering
    \includegraphics[scale=0.67]{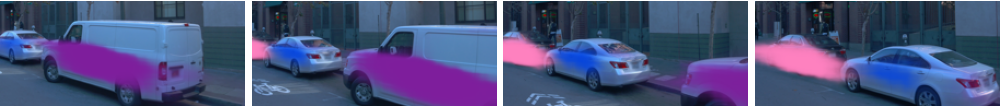}
    \caption{Our model is able to track many vehicles and pedestrians in held-out validation runs from the real-world Waymo Open dataset. We use Hungarian matching to determine best-fit predicted masks for each ground-truth object as described in section \ref{sec:quantitative_results}. In this figure, the alpha-value of the colored mask pixels is determined by the weight of the mask at each pixel.}
    \label{fig:tracking}
\end{figure*}

\begin{figure*}
    \centering
    \includegraphics[scale=0.67]{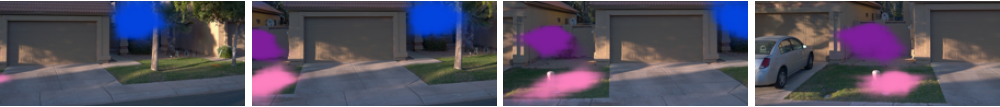}
    \caption{Due to its partitioning of the scene, our model tracks background features as well as dynamic objects. Here, three of the masks from a turning sequence are shown.}
    \label{fig:bg_tracking}
\end{figure*}

We find that our model emergently learns to assign the same mask to the same parts of the scene across time and different camera viewpoints, such that the movement of the masks across time tends to track both dynamic vehicles and pedestrians and static background features in the scene. An example of the learned masks for all three camera viewpoints is shown in figure \ref{fig:fusion}. Although the learned masks do not always provide a good shape match for the tracked objects, they do tend to follow them consistently across multiple time-steps and camera viewpoints. Examples of tracking vehicles and background features are shown in figures \ref{fig:tracking} and \ref{fig:bg_tracking}, respectively. 

Because we provide no supervision for which parts of the scene the object slots should attend to, the masks provide an exhaustive partition of the scene. We note that many elements of the background in this complex real-world dataset don't lend themselves to clear object-centric interpretations, so the correct partition is ambiguous.

\begin{figure}
    \centering
    \includegraphics[scale=0.53]{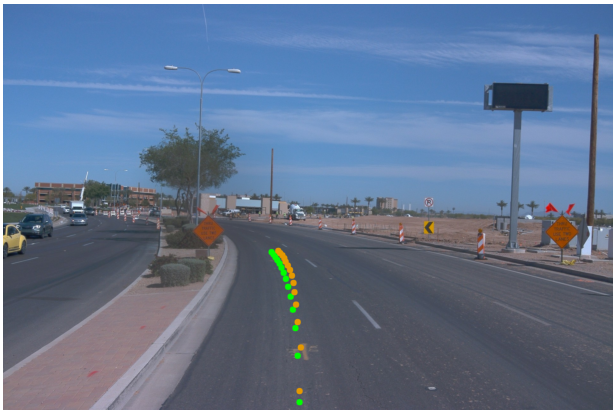}
    \caption{An example of the ground-truth future waypoints (green) and waypoints predicted by our model (orange). The predictions are for 1.6s into the future after the last frame in the input sequence.}
    \label{fig:waypoints}
\end{figure}

We additionally find that the waypoint prediction auxiliary task learns to predict qualitatively reasonable trajectories, as shown in figure \ref{fig:waypoints}. This suggests that the object-centric representation is compatible with learning to predict appropriate driving actions.

\subsection{Quantitative results}
\label{sec:quantitative_results}
We compare our model's performance with the prior state of the art models \cite{kipf_conditional_2022} in table \ref{tbl:metrics}. The SAVi++ models receive additional depth supervision, and the conditional version of SAVi++ is additionally initialized with bounding boxes around the ground-truth objects in the first frame of the sequence. The SAVi (RGB) model, like ours, does not receive depth or bounding box supervision.

We report sequence Adjusted Rand Index for Foreground objects (ARI-F) and centroid tracking. ARI-F is calculated for all instance labels in the sequence, which for the Waymo Open dataset include vehicles and pedestrians, but ignores background objects such as buildings or signs that do not typically exhibit independent motion. To measure tracking performance, we use the Center-of-Mass (CoM) distance metric proposed by \cite{kipf_conditional_2022}. The centroids of ground-truth instances are calculated via the geometric mean of their masks in each frame, and for the predicted slots via the weighted geometric mean of each slot's $\alpha$ mask. Ground-truth instances that make up more than 0.05\% of the total per-camera pixels are matched with the closest predicted centroid via Hungarian matching. We report the distance normalized by the length of the frame diagonal and averaged across frames and instances.

While our model does not achieve as good center-of-mass tracking as SAVi++, it approaches its performance without depth supervision. Our model also significantly outperforms vanilla SAVi while receiving only camera pose as an additional input. 

Furthermore, our results suggest that learning to predict the future pose of the ego vehicle is a synergistic task that does not harm the object-centric scene decomposition as measured by center-of-mass tracking performance. When training data is collected from expert drivers, future waypoint prediction (also known as behavioral cloning) is a leading method for picking driving actions in an autonomous driving context \cite{chen_learning_2022}. Our results suggest that the object-centric representation learned by our model may be usable directly as a vision backbone for AVs.

\begin{table*}[h!]
\caption{Object segmentation performance metrics on the Waymo Open dataset. SAVi, SAVi++, and depth-augmented SIMONe metrics are those reported in \cite{elsayed2022savi++} and may differ slightly in data-loading methodology. We report the standard error across 3 random seeds.}
\centering
\begin{tabular}{ |c|c|c|c| } 
\label{tbl:metrics}
Model & Privileged Information & ARI-F $\uparrow$ & CoM (\%) $\downarrow$ \\
\hline
SAVi (RGB) & - & - & 21.5 $\pm$ 1.8 \\
SAVi++ & Bounding boxes and depth & - & 4.4 $\pm$ 0.2 \\
SAVi++ (unconditioned) & Depth & - & 6.9 $\pm$ 0.5\\
SIMONe & Depth & - & 7.4 $\pm$ 0.2 \\
\hline
Ours (no Gaussian mixture) & Camera pose & 0.193 $\pm$ 0.004 & 10.0 $\pm$ 0.3 \\
Ours (no viewpoint) & None & 0.237 $\pm$ 0.003 & 9.8 $\pm$ 0.3 \\
Ours (no waypoint pred.) & Camera pose & 0.257 $\pm$ 0.018 & 9.9 $\pm$ 0.7 \\
\textbf{Ours} & \textbf{Camera pose} & \textbf{0.253 $\pm$ 0.009} & \textbf{9.6 $\pm$ 0.4} \\
\end{tabular}
\end{table*}

\subsection{Ablations}
In table \ref{tbl:metrics} we also report metrics from ablations of our model. In the no Gaussian mixture condition, the number of Gaussian modes for the pixel reconstruction distribution is $H$=1. In the no viewpoint condition, the transformer patch positional embeddings and decoder pixel coordinate queries refer only to the index of the camera (left, front, or right) and position in the sequence rather than containing the camera transformation matrix. Note that this is not equivalent to the original SIMONe model because we do not infer separate ``frame latents" to represent the camera viewpoint. In the no waypoint prediction condition, we do not add the waypoint prediction auxiliary task. Hyperparameters are kept constant across models, except for the waypoint prediction condition which uses a slightly higher $\beta$ (5e-7 vs. 4.5e-7) to ensure that the KL-divergence of the object latent distribution is similar to the other models.

We find that both removing the viewpoint supervision and the Gaussian mixture decoding mixture significantly harm the ARI-F of the model, and there is additionally a trend towards slightly higher CoM in these conditions. The waypoint prediction auxiliary task has little effect on the segmentation metrics, suggesting that the object-centric representation can be used in end-to-end learning with useful downstream tasks.

\subsection{Limitations}
While the fusion and tracking properties of our model suggest that it is learning to capture the movement of objects in the scene, the quality of the masks still lags behind what is possible with supervised instance segmentation \cite{he2017mask,kirillov2023segment} and what has been demonstrated with self-supervised learning on synthetic datasets \cite{kabra_simone_2021}. Two failure modes exhibited by our model are ``over-segmentation," where more than one mask is used to represent the same object, and mask re-use, where the same slot represents multiple independent parts of the scene (as can be seen in figure \ref{fig:fusion}). We found that simply increasing the number of slots available to the model does not resolve mask re-use and may lead to increased tessellation-type failures, an issue that has been reported in other object-centric models \cite{sajjadi2022object,karazija2021clevrtex}. While in the self-supervised setting the correct partitioning of the scene is ambiguous, both of these limitations could be concerning for the use of object-centric representations in downstream driving tasks such as planning and scene understanding, and therefore merit further investigation.

\section{Conclusion}
Self-supervised object-centric representation learning approaches have recently shown strong performance on synthetic datasets with clearly defined objects, but continue to struggle with complex real-world data with complicated textures and ambiguous objects. In this paper we have presented results suggesting that by using camera pose as additional input, it is possible to obtain reasonable dynamic, object-centric representations in the context of RGB driving video. Since pose estimation, unlike 3D depth sensors, is a ubiquitous feature of autonomous vehicles, we believe that our approach is a promising avenue towards scalable and practical object-centric representation learning in the context of autonomous driving. Furthermore, our results suggest that predicting the future pose of the ego vehicle is a synergistic task that doesn't hinder the quality of the learned representations. This is particularly exciting for end-to-end driving models because it opens up the possibility of driving performance and representation learning improving together in a virtuous cycle while retaining key advantages of object-centric representations, such as interpretability.

We suspect further improvements to the object segmentation quality are still possible, for example via careful scaling up of model size and a data augmentation strategy (both of which were important for the performance of SAVi++ \cite{kipf_conditional_2022}). We also note that the size of the Waymo Open perception dataset, containing 480,000 frames across the three forward-facing cameras, may not be sufficiently large relative to its complexity for ideal representation learning. The recent Object Scene Representation Transformer model, in contrast, was trained on 10M frames of a synthetic dataset \cite{sajjadi2022object}. 

Finally, we note that the KL-divergence loss in our model encourages learning disentangled object latent features \cite{higgins2017beta}. Examining these features in more detail is an exciting direction for future work.

\section{Appendix}
\subsection{Additional architecture details}

Hyperparameters for our model are shown in table \ref{tbl:hyperparams}. The metrics in table \ref{tbl:metrics} are reported after 2e5 training steps. We found that minor improvements in segmentation quality continued after this point, and qualitative figures in the paper are from a model trained for up to 6e5 steps. The transformer used for decoding in the waypoint prediction task used the same hyperparamters as the main model transformer, except with 2 layers instead of 6.

\begin{table}[h!]
\caption{Model hyperparameters used in our experiments.}
\centering
\begin{tabular}{ |l|l| } 
\label{tbl:hyperparams}
Parameter & Value \\
\hline
$\mathcal{L}_{KL}$ weight, $\beta$ (with waypoint pred. task) & 5e-7 \\
$\beta$ (without waypoint pred. task) & 4.5e-7 \\
$\mathcal{L}_{task}$ weight, $\omega_{task}$ & 1e-4 \\
Pixel likelihood scale, $\sigma_x$ & 0.08 \\
Number of object slots & 21 \\
Object latent dim. & 32 \\
Transformer num. layers & 6 \\
Transformer num. heads & 4 \\
Transformer feature dim. & 512 \\
Transformer feedforward dim. & 1024 \\
Reconstruction MLP num. layers & 3 \\
Reconstruction MLP hidden dim. & 1536 \\
Image sequence length & 8 \\
Image size (H, W) & (96, 224) \\
Pixels decoded per training sequence, $N$ & 2016 \\
Batch size & 8 \\
Learning rate & 1e-4 \\
\end{tabular}
\end{table}

{\small
\bibliographystyle{ieee_fullname}
\bibliography{main}
}

\end{document}